\documentclass[letterpaper, 10 pt, conference]{ieeeconf}  % Comment this line out if you need a4paper

\IEEEoverridecommandlockouts                              % This command is only needed if
\usepackage{epsfig}
\usepackage{graphicx}
\usepackage{amsmath}
\usepackage{amssymb}
\usepackage{subcaption}
\usepackage{url}
\usepackage{textcomp}
\usepackage{upgreek}
\usepackage{bm}
\usepackage{cite}
\usepackage{multirow}
\usepackage{rotating}
\usepackage{booktabs}
\usepackage{listings}

\title{\LARGE \bf
Pushbroom Stereo for High-Speed Navigation in Cluttered Environments
}

\author{Andrew J. Barry and Russ Tedrake% <-this % stops a space
\thanks{The authors are with the Computer Science and Artificial Intellegence Laboratory, Massachusetts Institute of Technology, Cambridge, MA, USA.
        {\tt\small \{abarry, russt\}@csail.mit.edu}}% <-this % stops a space
\thanks{This work was supported by ONR MURI grant N00014-09-1-1051. Andrew Barry is partially supported by a National Science Foundation Graduate Research Fellowship.}%
}

\begin{document}

\maketitle
\thispagestyle{empty}
\pagestyle{empty}

%%%%%%%%%%%%%%%%%%%%%%%%%%%%%%%%%%%%%%%%%%%%%%%%%%%%%%%%%%%%%%%%%%%%%%%%%%%%%%%%
\begin{abstract}

We present a novel stereo vision algorithm that is capable of obstacle detection on a mobile-CPU processor at
120 frames per second.  Our system performs a subset of standard block-matching stereo processing, searching
only for obstacles at a single depth.  By using an onboard IMU and state-estimator, we can recover the
position of obstacles at all other depths, building and updating a full depth-map at framerate.

Here, we describe both the algorithm and our implementation on a high-speed, small UAV, flying at over 20
MPH (9 m/s) close to obstacles.  The system requires no external sensing or computation and is, to the best
of our knowledge, the first high-framerate stereo detection system running onboard a small UAV.

\end{abstract}

%%%%%%%%%%%%%%%%%%%%%%%%%%%%%%%%%%%%%%%%%%%%%%%%%%%%%%%%%%%%%%%%%%%%%%%%%%%%%%%%
\section{Introduction}
%   UAVs need to avoid things
%       but sensors are heavy
%       so use cameras
%           but that's too slow

Recently we have seen an explosion of exciting results on small, unmanned aerial vehicles (UAVs) such as
incredible obstacle avoidance and trajectory tracking \cite{Mellinger10}, formation flight
\cite{Mellinger10c, Kushleyev12}, and cooperative interactions with the environment \cite{Ritz12, Lindsey12,
Brescianini13}.  All these systems, however, rely on an external motion-capture apparatus that gives the
vehicles almost perfect state information at high rates.  As we move these tasks out of the lab and into the
field, we need new techniques to provide this sensing information.

A major challenge in gathering sensing data necessary for flight is the limited payload, computation, and
battery life of the vehicles.  These small aircraft, weighing under 1-2kg, struggle to carry enough
sensing payload for high-speed navigation in complex 3D environments.  Lightweight cameras are a good
solution, but require computationally efficient machine vision algorithms that can run within the limits
of these vehicles.  For these reasons, camera based systems have so far been limited to relatively slow,
stable flight regimes \cite {Engel12, Sa13}.  We aim to fly at speeds of 7-15m/s through cluttered
environments like a forest, well outside the typical speed regime. The short wingspan required to fit
through gaps in the clutter limits our payload capacity while we need fast framerates with short exposures
to avoid motion blur and to have enough time to avoid oncoming obstacles.

To this end, we propose a novel method for stereo vision computation that is dramatically faster than the state of
the art.  Our method performs a subset of the processing traditionally required for stereo vision, but is
able to recover obstacles in real time at 120 frames-per-second (fps) on a conventional CPU.  Our system is
lightweight and accurate enough to run in real time on our aircraft, allowing for true, self-contained
obstacle detection.

\begin{figure}
\begin{center}
\includegraphics[width=\columnwidth]{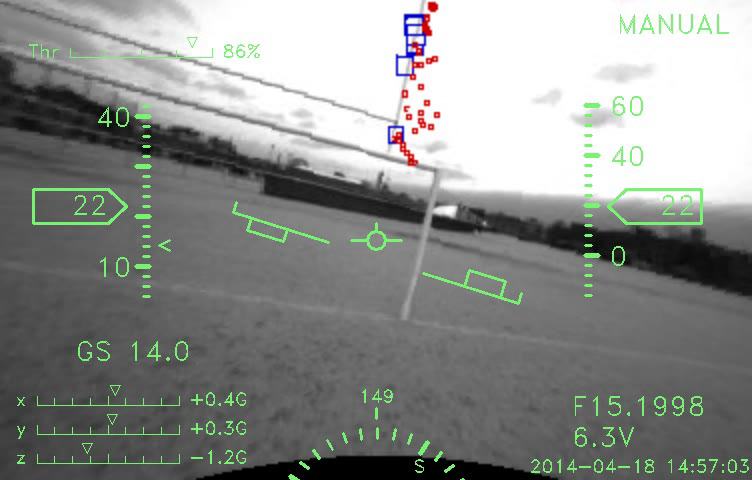}
\caption{
    \label{inFlight}
    In-flight snapshot of single-disparity stereo detections on a goalpost (blue boxes) and past detections
    integrated through the state estimate and reprojected back on the image (red dots).  Overlay includes
    relevant flight data such as airspeed in MPH (left) and altitude in feet (right).
}
\end{center}
\end{figure}

\section{Related work}
\subsection{Obstacle Detection on UAVs}

Obstacle detection on small outdoor UAVs continues to be a challenging problem.  Laser rangefinders usually
only support 2D detections and are generally too heavy for flight, although some systems with large wingspans
\cite{Bry12} or limited flight time \cite{Richter13} exist. Other active rangefinders such as the Microsoft
Kinect\footnote{Microsoft, Inc. \url {http://www.microsoft.com/en-us/kinectforwindows/}} and
PrimeSense\footnote{PrimeSense, LTD. \url {http://www.primesense.com/}} systems rely on less focused infrared
light and do not work in bright outdoor environments.  Here we detail some of the related vision and other
lightweight sensors for this class of vehicles.

\subsection{Optical Flow}

Embedded optical flow techniques rely on hardware (such as commonly found in optical mice) to compute the
inter-frame changes between images to extract depth information.  These techniques have worked well on UAVs,
demonstrating autonomous takeoff, landing \cite{Beyeler09a, Barber07} and obstacle avoidance
\cite{Zufferey08, Beyeler09}.  This technology has been successful for aircraft flight control and is now
available commercially\footnote{senseFly LTD. \url{http://www.sensefly.com/}}.  Embedded optical
flow, however, is limited in its resolution, providing only general guidence about obstacles.  For more
sophisticated flight, such as flying in a cluttered environment like a forest, we must look beyond optical
flow techniques for solutions that provide greater resolution.

\subsection{Monocular Vision}

Monocular vision techniques are attractive because they rely on only a single, lightweight camera, are readily
available, and easy to deploy.  State of the art monocular depth estimation, however, is generally not fast and reliable
enough for obstacle avoidance on fast-flying UAVs.  Machine learning work has shown progress, such as Michels's
radio controlled car \cite{Michels05}.  More modern work includes scale-estimation in hover and slow flight
regimes \cite{Sa13}. %PTAM + AR drone + peter corke = decent hover
When integrating pressure and inertial measurement sensors such as Achtelik et al.'s implemention of PTAM
(parallel tracking and mapping)
\cite{Klein07} with a barometric altimeter, stable flights in indoor and outdoor environments are possible
\cite{Achtelik11a}.  With a full vison-aided inertial navigation system (VINS), Li et al.\ have shown
remarkable tracking with commodity phone hardware, demonstrating tracking within 0.5-0.8\% of distance
traveled for significant distances \cite{Li13, Li13a}.

% todo: Shen13, Shen13a

\subsection{Stereo Vision}

While we have seen substantial improvements in monocular vision systems recently, they are not yet fast or accurate
enough for high-speed obstacle avoidance on small UAVs.  Stereo systems suffer from a similar speed issue, with most
modern systems running at or below 30 Hz \cite{Yang03, Byrne06}.

Honegger et al.\ recently demonstrated an FPGA (Field Programmable Gate Array) based system that can compute
optical flow and depth from stereo on a 376x240 image pair at 127 fps or 752x480 at 60 fps \cite{Honegger12,
Honegger14}.  Their system is small and lightweight enough for use on a small UAV, but requires specialized
hardware and has not yet been flight tested.  By comparison, our approach performs less computation and can
work easily on conventional hardware, but relies on the observation that it is sufficient for high-speed
flight to compute a subset of the stereo matches.

\section{Proposed Method}

\begin{figure}
\centering
\includegraphics[width=\columnwidth]{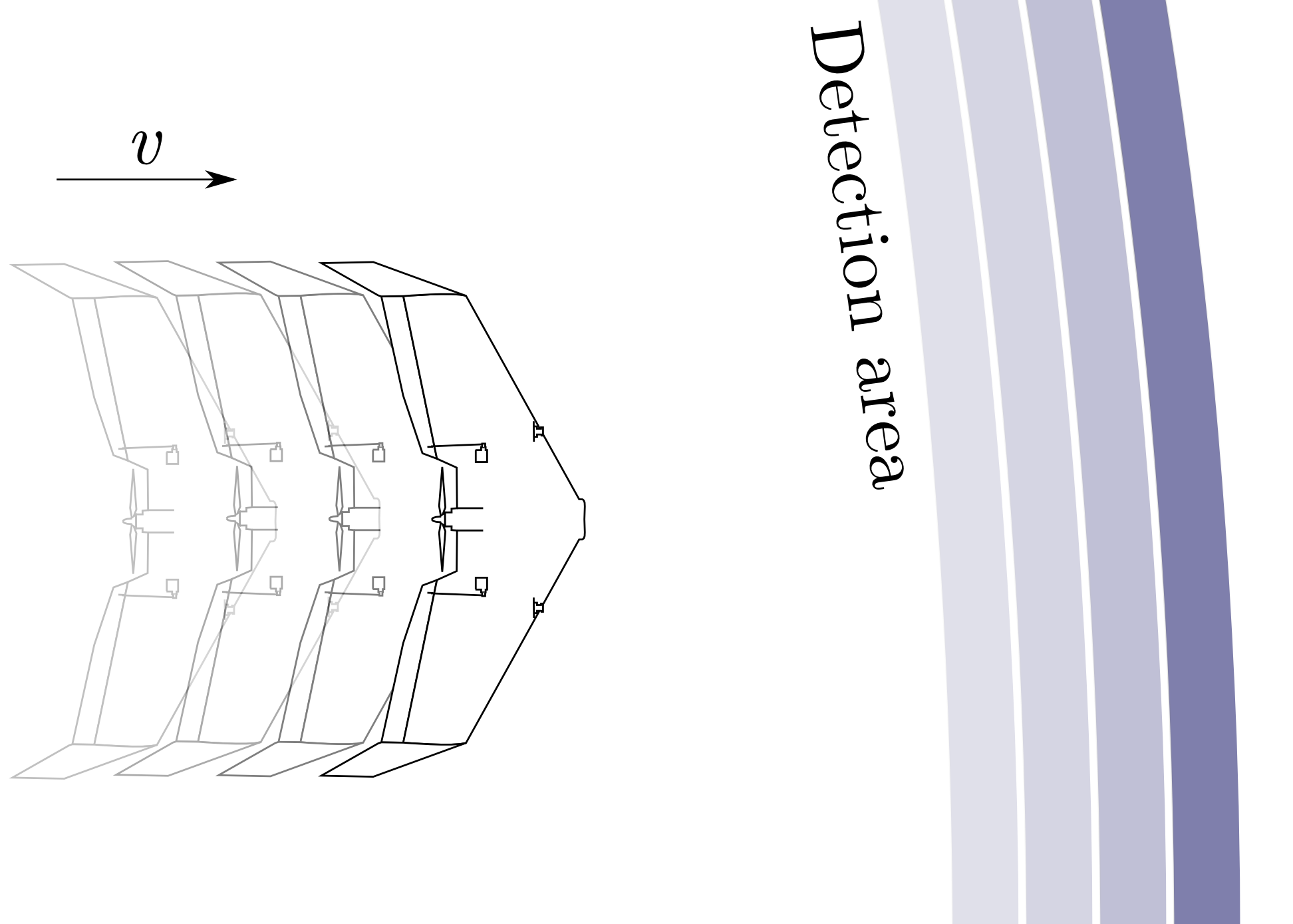}
\caption{
    \label{movingDetection}
    By detecting at a single depth (dark blue) and integrating the aircraft's odometry and past detections
    (lighter blue), we can quickly build a full map of obstacles in front of our vehicle.
}
\end{figure}

\subsection{Block-Matching Stereo}

A standard block-matching stereo system produces depth estimates by finding pixel-block matches between two
images.  Given a pixel block in the left image, for example, the system will search through the epipolar
line\footnote{Standard calibration and rectification techniques provide a line, called the epipolar line, on
which the matching block is guaranteed to appear.} to find the best match.  The position of the match
relative to its coordinate on the left image, or the disparity, allows the user to compute the 3D position of
the object in that pixel block.

\subsection{Pushbroom Stereo}

One can think of a standard block-matching stereo vision system as a search through depth.  As we search
along the epipolar line for a pixel group that matches our candidate block, we are exploring the space of
distance away from the cameras.  For example, given a pixel block in a left image, we might start searching
through the right image with a large disparity, corresponding to an object close to the cameras.  As we
decrease disparity (changing where in the right image we are searching), we examine pixel blocks that
correspond to objects further and further away, until reaching zero disparity, where the stereo base distance
is insignificant compared to the distance away and we can no longer determine the obstacle's location.

Given that framework, it is easy to see that if we limit our search through distance to a single value, $d$
meters away, we can substantially speed up our processing, at the cost of neglecting obstacles at distances
other than $d$. While this might seem limiting, our cameras are on a moving platform (in this case, an
aircraft), so we can quickly recover the missing depth information by integrating our odometry and previous
single-disparity results (Figure \ref{movingDetection}).  The main thing we sacrifice is the ability to take
the best-matching block as our stereo match; instead we must threshold for a possible match.

We give this algorithm the name ``pushbroom stereo" because we are ``pushing" the detection region forward,
sweeping up obstacles like a broom on a floor (and similar to pushbroom LIDAR systems \cite{Napier13}). We note
that this is distinct from a ``pushbroom camera," which is a one-dimensional array of pixels arranged
perpendicular to the camera's motion \cite{Gupta97}.  These cameras are often found on satellites and
can be used for stereo vision \cite{Hirschmuller05}.

\subsection{Odometry}

Our system requires relatively accurate odometry over short time horizons.  This requirement is not
particularly onerous because we do not require long-term accuracy like many map-making algorithms.  In our
case, the odometry is only used until the aircraft catches up to its detection horizon, which on many
platforms is 5-10 meters away.  We demonstrate that on aircraft, a wind-corrected airspeed measurement (from
a pitot tube) is sufficient.  On a ground robot, we expect that wheel odometry would be adequate.

%To generate wind-corrected measurements, we use an onboard GPS and barometric altimeter to measure the difference between airspeed and
%ground speed.  With time, we can estimate the average wind in three dimensions.  While we are unable to measure gusts
%beyond what our accelerometers and gyroscopes can integrate, this has proven sufficient for high-speed obstacle
%avoidance. % TODO: make sure the final system actually implements all of this

\begin{figure}
\begin{center}
\includegraphics[width=\columnwidth]{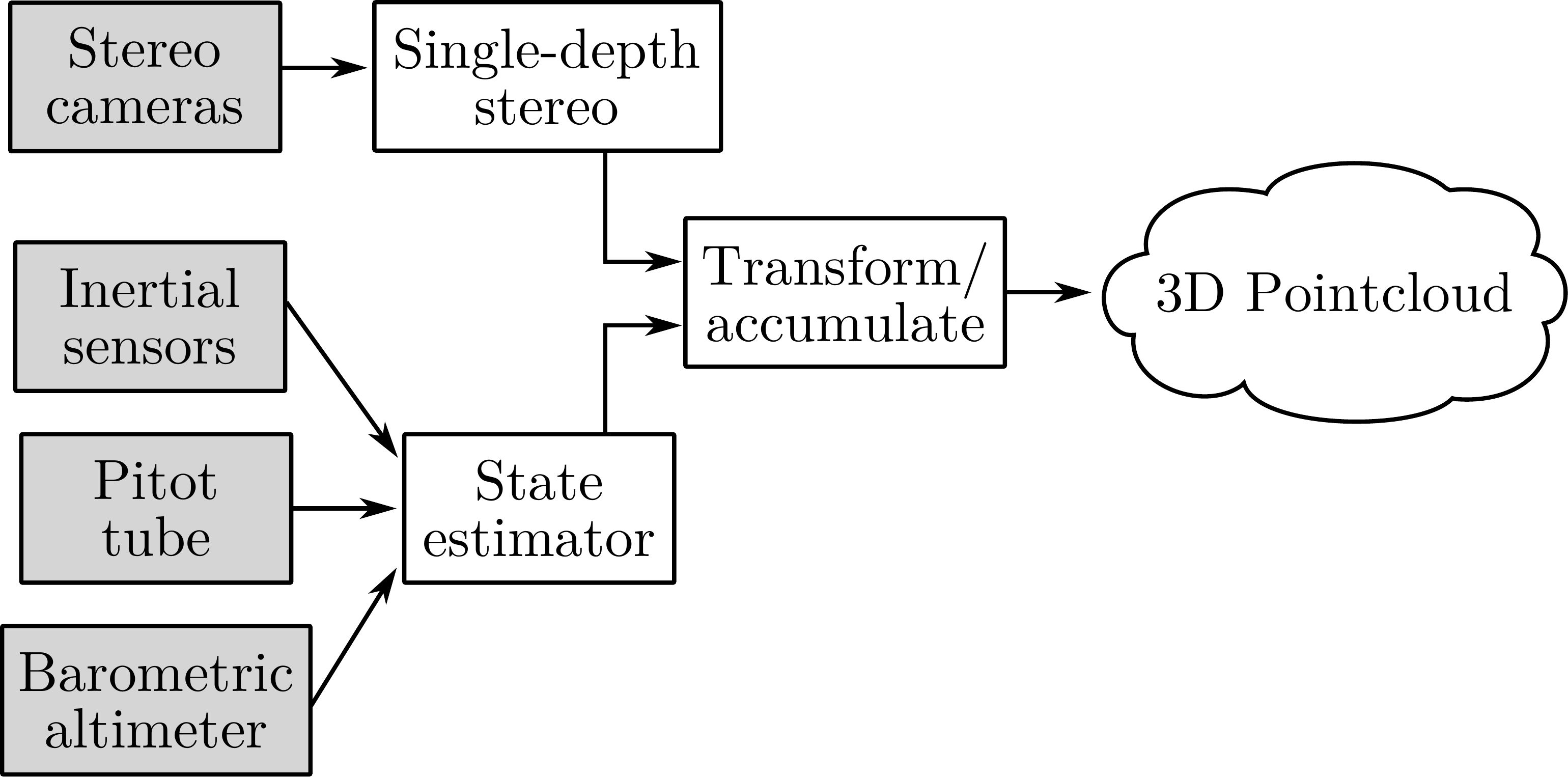}
\caption{
    \label{blockDiagram}
    Pushbroom stereo overview.
}
\end{center}
\end{figure}

\section{Implementation}

\subsection{Pushbroom Algorithm}

Like other block-matching algorithms, we use sum of absolute differences (SAD) to detect pixel block
similarity.  In addition to detecting matching regions, we score blocks based on their abundance of edges.
This allows us to disambiguate the situation where two pixel blocks might both be completely black, giving a
good similarity score, but still not providing a valid stereo match.  To generate an edge map, we use a
Laplacian with an aperture size (\texttt{ksize}) of 3.  We then take the summation of the 5x5 block in the
edge map and reject any blocks below a threshold for edge-abundance.

After rejecting blocks for lack of edges, we score the remaining blocks based on SAD match divided by the
summation of edge-values in the pixel block:
\[
S = \frac{
    \overbrace{
        \sum\limits_{i=0}^{5\text{x}5}
            \lvert p(i)_{left} - p(i)_{right} \rvert
    }^{
        \text{Sum of absolute differences (SAD)}
    }
}
{\sum\limits_{i=0}^{5\text{x}5}{L \left(p(i)_{left} \right) + L \left(p(i)_{right} \right) }}
\]
where $p(i)$ denotes a pixel value in the 5x5 block and $L$ is the Laplacian.  We then threshold on the
score, $S$, to determine if there is a match.

We have deliberately chosen a design and parameters to cause sparse detections with few false positives.  For
obstacle avoidance, we do not need to see every point on an obstacle but a false positive might cause the
aircraft to take unnecessary risks to avoid a phantom obstacle.

\begin{figure}
    \centering
    \begin{subfigure}[b]{0.48\columnwidth}
        \includegraphics[width=\textwidth]{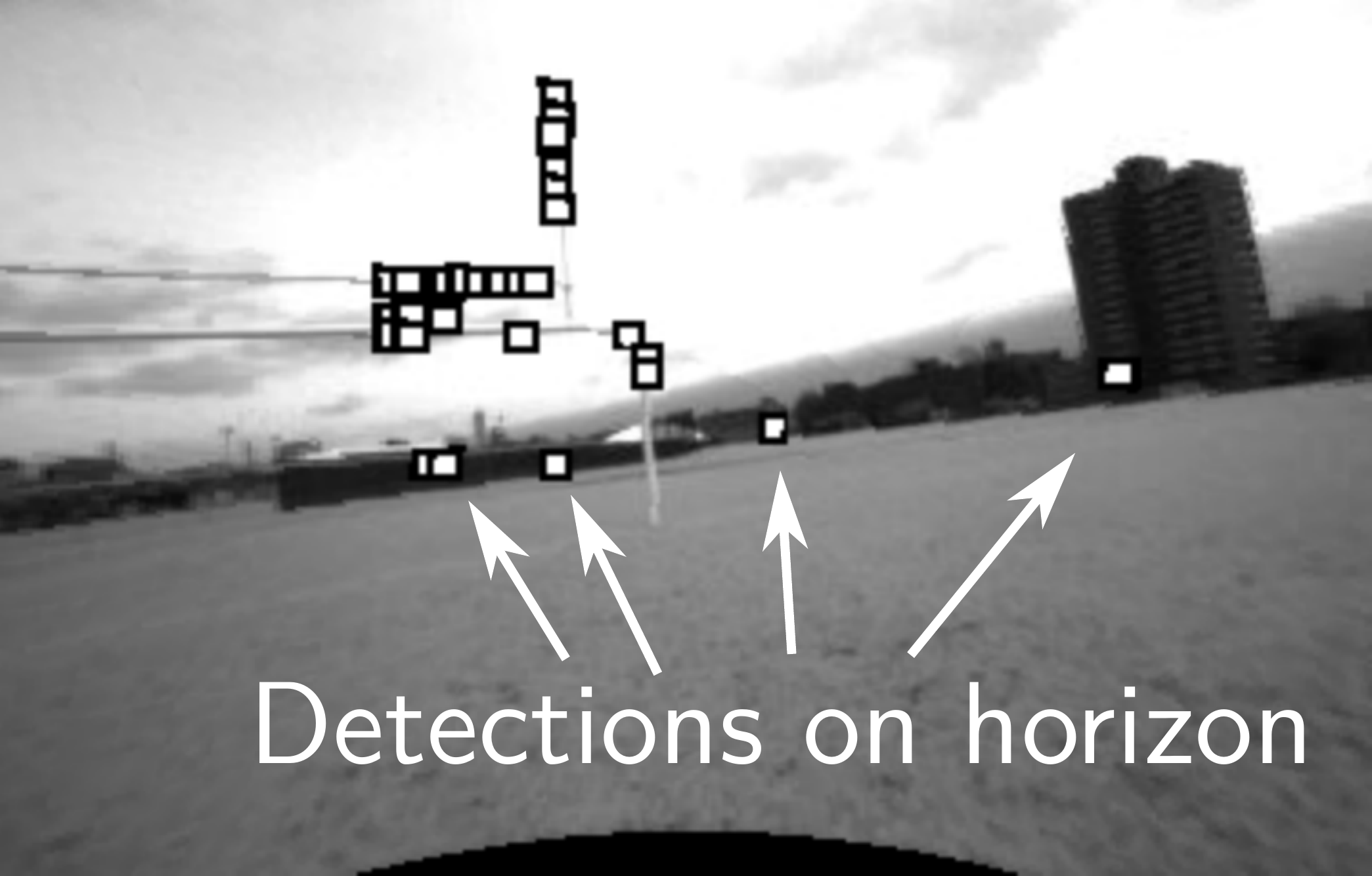}
        \caption{Without horizontal invariance filter.}
    \end{subfigure}
    \hfill
    \begin{subfigure}[b]{0.48\columnwidth}
        \includegraphics[width=\textwidth]{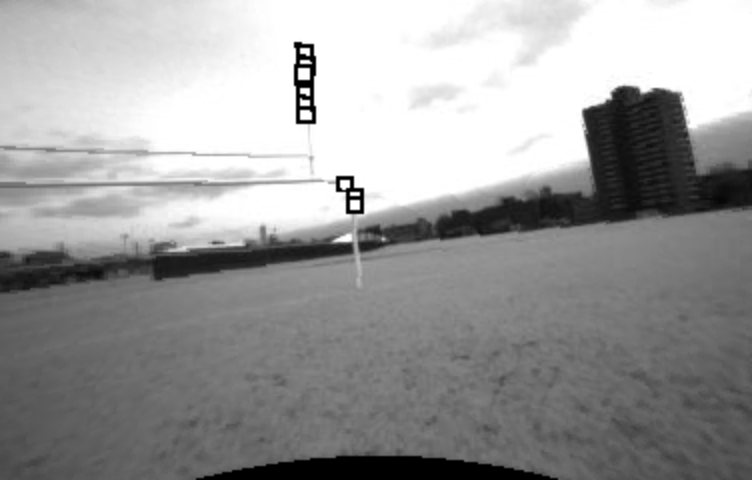}
        \caption{Horizontal invariance filter enabled.}
    \end{subfigure}

    \caption{
        All stereo systems suffer from repeating textures which cannot be disambiguated with only two
        cameras.  Here, we demonstrate our filter for removing self-similarity.  Detected pixel blocks
        are marked with squares.  Note that the filter removes all self-similar regions including those on
        obstacles, limiting our ability to detect untextured, horizontal obstacles.
        \label{horizontalInvariance}
    }

\end{figure}

All two-camera stereo systems suffer from some ambiguities.  With horizontal cameras, we cannot disambiguate
scenes with horizontal self-similarity, such as buildings with grid-like windows or an uninterrupted horizon.
These horizontal repeating patterns can fool stereo into thinking that it has found an obstacle when it has
not.

While we cannot correct these blocks without more sophisticated processing or additional cameras, we can
detect and eliminate them.  To do this, we perform additional block-matching searches in the right image
near our candidate obstacle.  If we find that one block in the left image matches blocks in the right
image at different disparities, we conclude that the pixel block exhibits local self-similarity and reject
it.  While this search may seem expensive, in practice the block-matching above reduces the search size so
dramatically that we can run this filter online.
 Figure \ref{horizontalInvariance} demonstrates this filter running on flight data.

\begin{figure}
\centering
\includegraphics[width=.8\columnwidth]{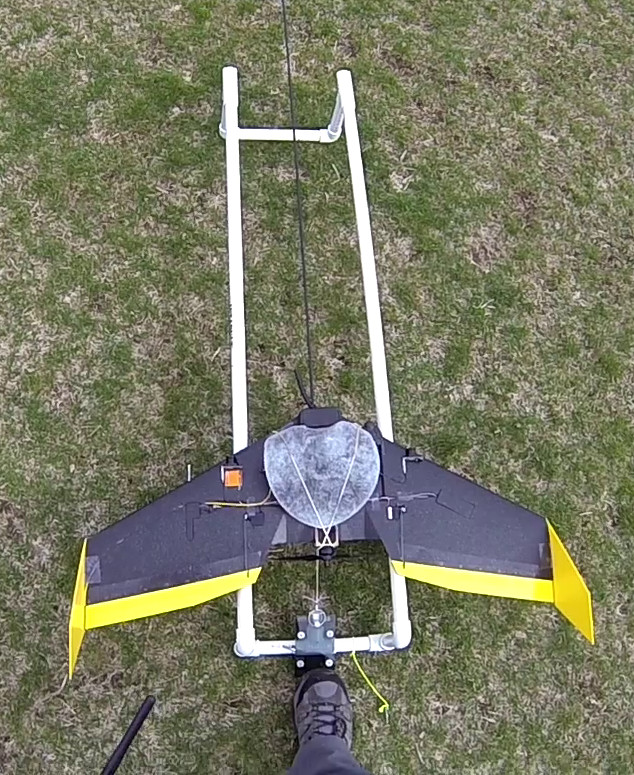} % TODO: update photo
\caption{
    \label{deltaPhoto}
    Aircraft hardware in the field.  We use a small catapult for consistent launches near obstacles.
}
\end{figure}

\subsection{Hardware Platform}

We implemented the pushbroom stereo algorithm on a quad-core 1.7Ghz ARM, commercially available in the
ODROID-U2 package, weighing under 50 grams\footnote{Hardkernel co., Ltd. \url{http://hardkernel.com}}.  Our
cameras' resolution and stereo baseline can support reliable detections out to approximately 5 meters, so we
use 4.8 meters as our single-disparity distance.  We detect over 5x5 pixel blocks, iterating through the left
image with 8 parallel threads.

We use two Point Grey Firefly MV\footnote{Point Grey Research, Inc. \url{http://www.ptgrey.com}} cameras,
configured for 8-bit grayscale with 2x2 pixel binning, running at 376x240 at 120 frames per second.  A second
ODROID-U2, communicating over LCM \cite{Huang10}, runs our state-estimatator (a 12-state Kalman filter from
\cite{Bry12}) and connects to our low-level interface, a firmware-modified APM 2.5\footnote{3D Robotics, Inc.
\url{http://3drobotics.com/}}, which provides access to our servo motors, barometric altimeter, pitot tube
airspeed sensor, and 3-axis accelerometer, gyroscope, and magnetometer suite.

This platform sits aboard a modified Team Black Sheep Caipirinha\footnote{Team Black Sheep,
\url{http://team-blacksheep.com/products/prod:tbscaipirinha}}, a 34 inch (86cm) wingspan delta wing with a 2000kV
brushless DC motor and 8-inch propeller\footnote{Graupner 8x5 carbon fiber propeller.} (Figure
\ref{deltaPhoto}).  All outdoor flights are conducted with the aircraft under control of a passively-spooling non-conductive 250 meter safety tether.

% what is the rest of this section going to look like?
%
% results for stereo system
%
%   obstacle detection with odometry
%       how do I show successful detections?
%           movies work but what do I do in a paper?
%
%           maybe a graph of true/false positives/negatives vs. offline stereo BM and offline global stereo?
%           also maybe something with online (failing at running fast) stereo?
%
%   comparison against full stereo when filtering for depth
%
%

\section{Results}

\subsection{Single-Disparity Stereo}\label{sec:singleDisparityResults}

\begin{figure}
\begin{center}
\includegraphics[width=.7\columnwidth]{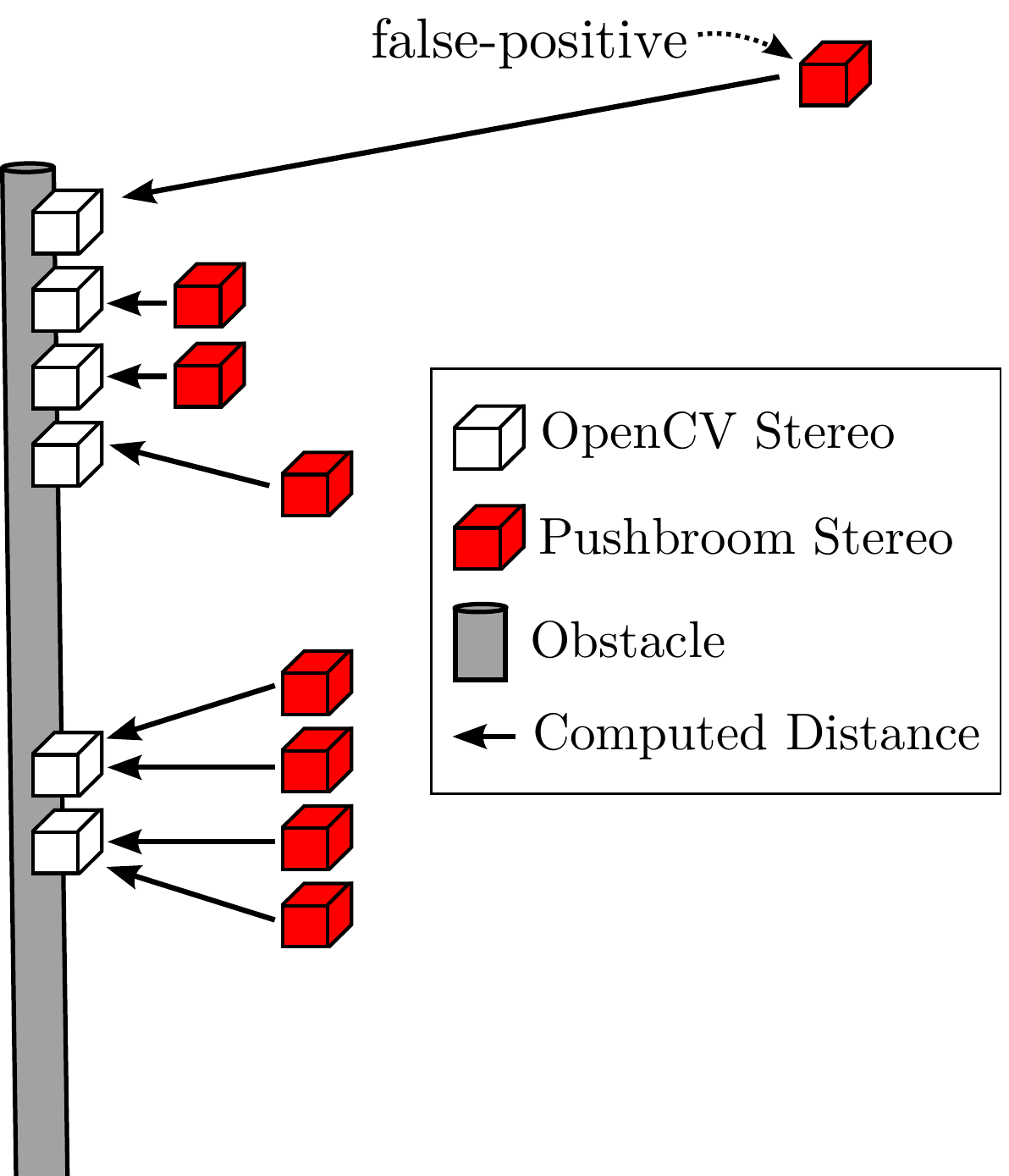}
\caption{
    \label{bmStereoSketch}
    Sketch of our evaluation strategy for single-disparity stereo.  We detect false-positives by computing the
    distance from single-disparity stereo's output (red) to the nearest point from OpenCV's \texttt{StereoBM}
    (white).  False positives stand out with large distances (labeled box).
}
\end{center}
\end{figure}

To determine the cost of thresholding stereo points instead of using the best-matching block from a search
through depth, we walked with our aircraft near obstacles and recorded the output of the onboard stereo
system with the state-estimator disabled\footnote{Our state-estimator relies on the pitot-tube airspeed
sensor for speed estimation, which does not perform well below flight speeds.}.  We then, offline, used
OpenCV's~\cite{Bradski00} block-matching stereo implementation (\texttt{StereoBM}) to compute a full depth
map at each frame.  We then removed any 3D point that did not correspond to a match within 0.5 meters of our
single-disparity depth to produce a comparison metric for the two systems.

With these data, we detected false-positives by computing the Euclidean distance from each single-disparity
stereo coordinate to the nearest point produced by the depth-cropped StereoBM (Figure \ref{bmStereoSketch}).
Single-disparity stereo points that are far away from any StereoBM points may be false-positives introduced
by our more limited computation technique.  StereoBM produces a large number of false negatives, so we do not
perform a false-negative comparison on this dataset (see Section \ref{sec:flightExperiments} below.)

\begin{figure}
\begin{center}
\includegraphics[width=\columnwidth]{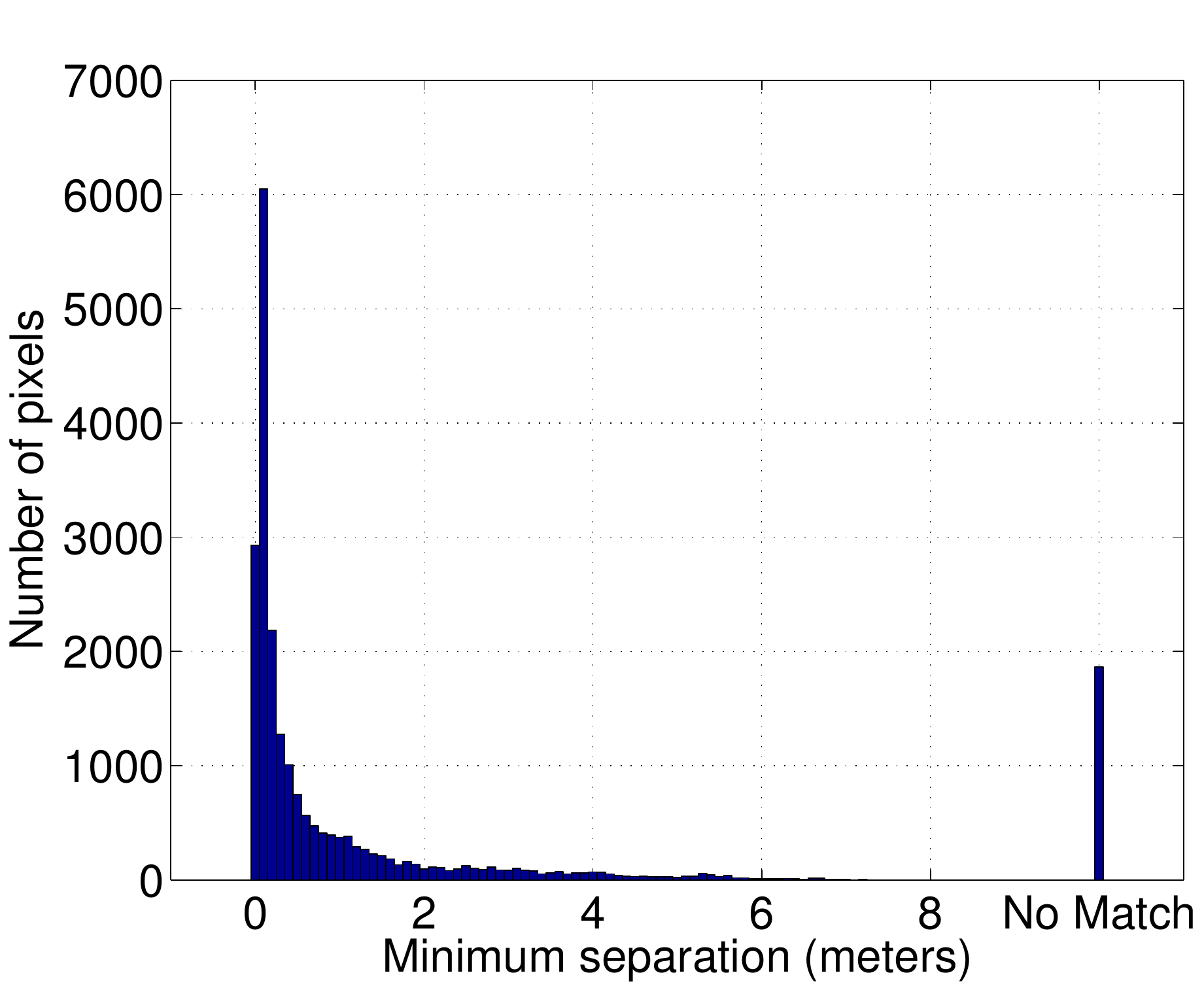}
\caption{
    \label{singleDepthStereo}
    Results of the false-positive benchmark described in Figure \ref{bmStereoSketch} on 23,000+ frames.
    \textit{No Match} indicates single-disparity points where there was no matching StereoBM point on the
    frame.  We find that only 8.2\% of detected pixels fall into this category.
}
\end{center}
\end{figure}

Our ground dataset includes over 23,000 frames in four different locations with varying lighting conditions,
types of obstacles, and obstacle density.  Over the entire dataset, we find that single-disparity stereo
produces points within 0.5 meters of StereoBM 60.9\% and within 1.0 meters 71.2\% of the time (Figure
\ref{singleDepthStereo}).  For context, the aircraft's wingspan is 0.86 meters and it covers 0.5 meters in
0.03 to 0.07 seconds.

\subsection{Flight Experiments}\label{sec:flightExperiments}

\begin{figure}
\centering
\hfill \includegraphics[width=.48\columnwidth]{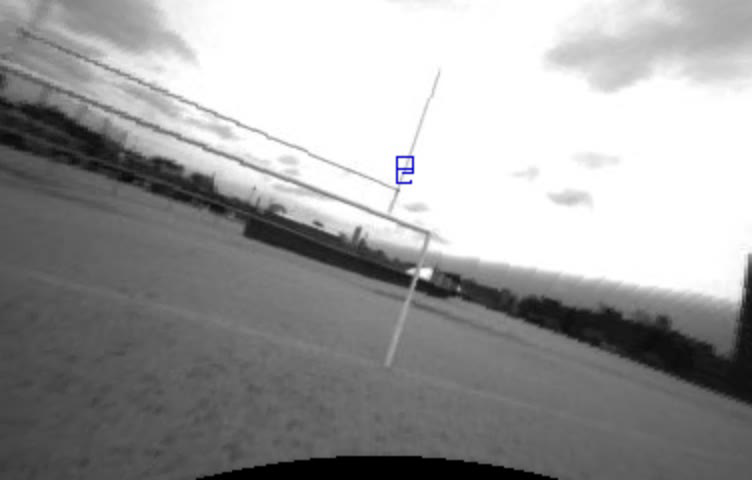}
\hfill \includegraphics[width=.48\columnwidth]{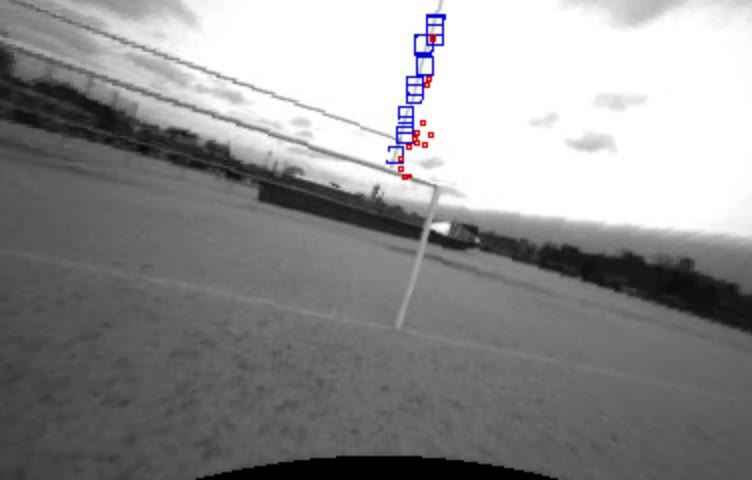} \hfill \\
\vspace{2pt}
\hfill \includegraphics[width=.48\columnwidth]{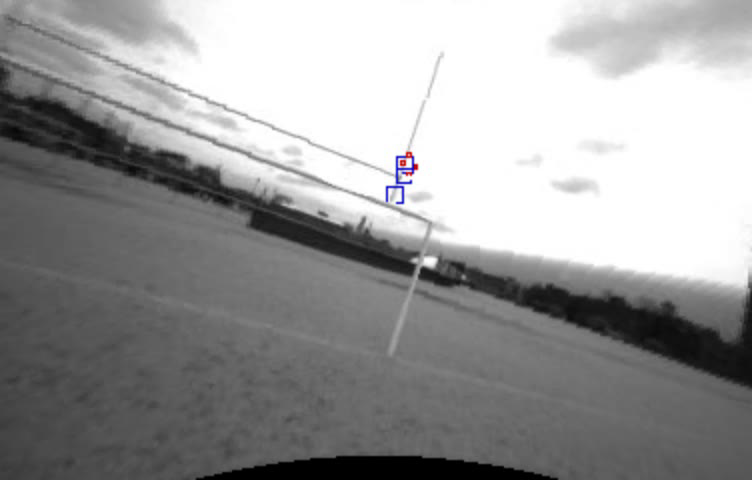}
\hfill \includegraphics[width=.48\columnwidth]{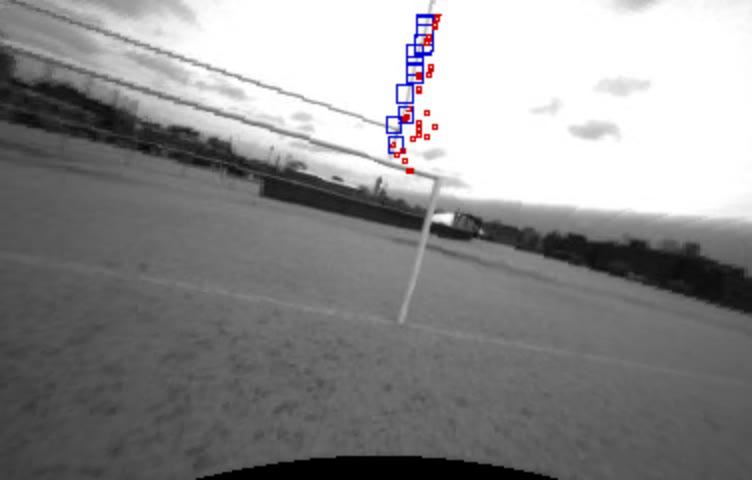} \hfill \\
\vspace{2pt}
\hfill \includegraphics[width=.48\columnwidth]{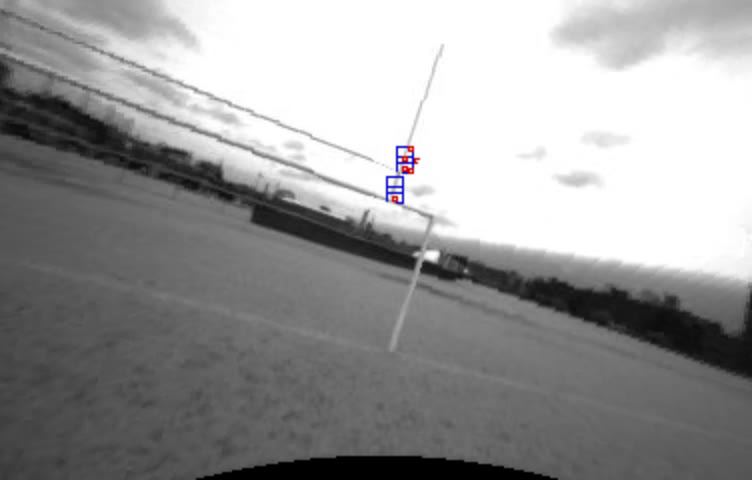}
\hfill \includegraphics[width=.48\columnwidth]{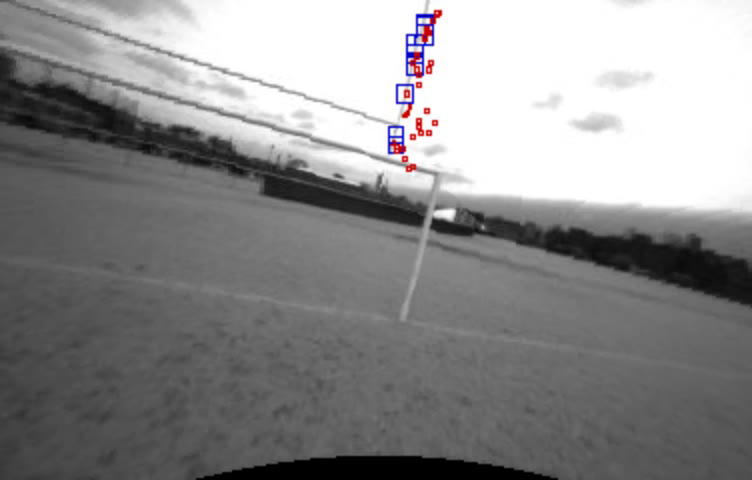} \hfill \\
\vspace{2pt}
\hfill \includegraphics[width=.48\columnwidth]{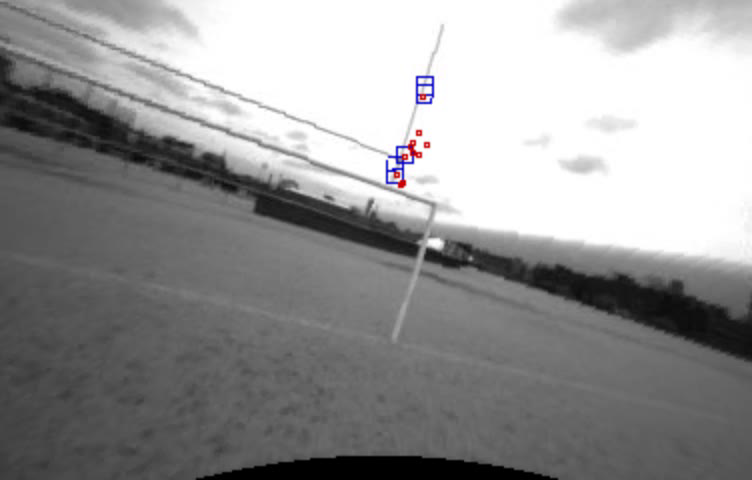} \hfill
       \includegraphics[width=.48\columnwidth]{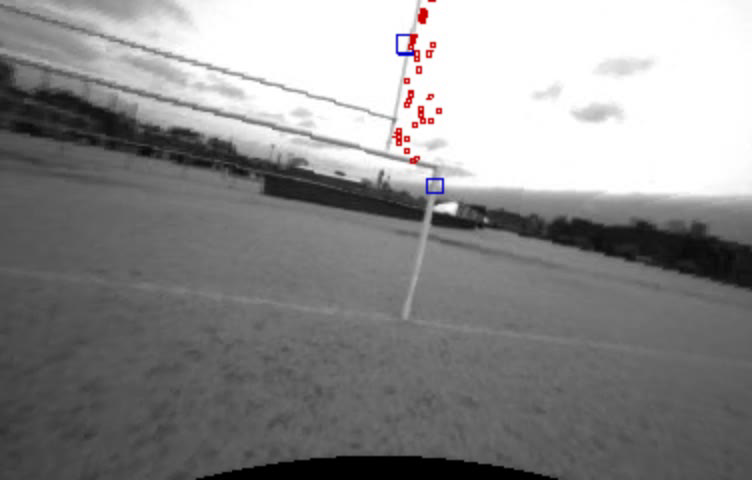} \\
\caption[Sequence of stills from an obstacle detection]{
    Sequence of stills from an obstacle detection.  Each image is 0.02 seconds (20ms) after the previous.
    The entire set captures 0.16 seconds.  Here, the fieldgoal is detected in the first frames (blue boxes).  Afterwards, the position of those detections is estimated via the state estimator and
    reprojected back onto the image (red dots).
    \label{stereoDetectionSeq}
}
\end{figure}

\begin{figure}
\begin{center}
\includegraphics[width=.35\columnwidth]{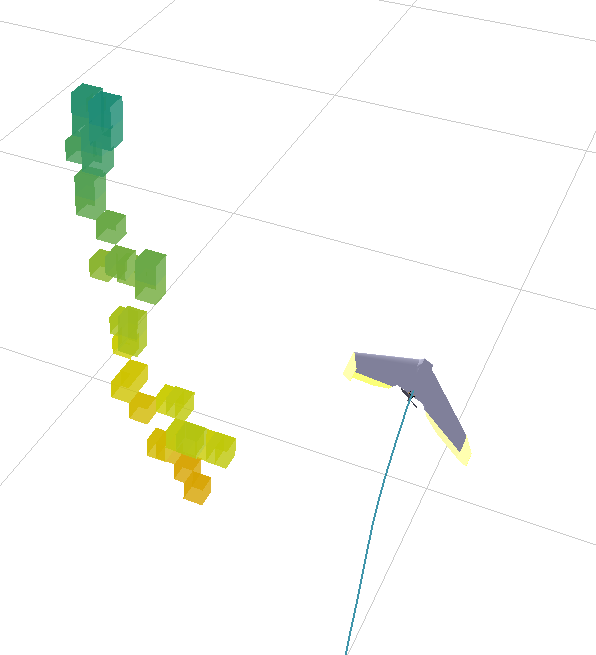}
\caption{
    \label{obstacleFlyby}
    Obstacle detection from Figure \ref{stereoDetectionSeq} rendered in a 3D visualizer.  While we do not
    endeavor to build maps, our system outputs pointclouds providing compatibility with many existing
    planning, visualization, and control tools.
}
\end{center}
\vspace{-5pt}
\end{figure}

To test the full system with an integrated state-estimator, we flew our platform close to obstacles
(Figure \ref{inFlight}) on three different flights, recorded control inputs, sensor data, camera images, and
on-board stereo processing results.  Figures \ref{stereoDetectionSeq} and \ref{obstacleFlyby} show on-board
stereo detections as the aircraft approaches an obstacle.

During each flight, we detected points on every obstacle in real time.  Our state estimate was robust enough
to provide online estimation of how the location of the obstacles evolved relative to the aircraft.  While
these flights were manually piloted, we are confident that the system could autonomously avoid the
obstacles with these data.  The integration of the planning and control system will be reported in future
work.

To benchmark our system, we again used OpenCV's block-matching stereo as a coarse, offline, approximation of
ground truth.  At each frame, we ran full block-matching stereo, recorded all 3D points detected, and then
hand-labeled regions in which there were obstacles to increase StereoBM's accuracy.

\begin{figure}
    \centering
    \begin{subfigure}[b]{0.24\columnwidth}
        \includegraphics[width=\textwidth]{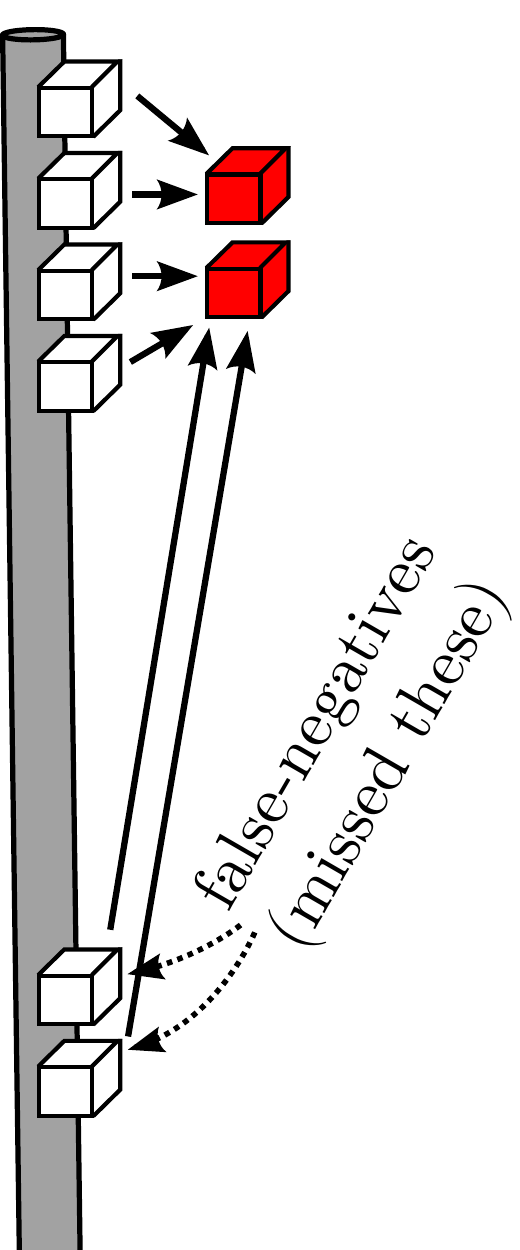}
        \caption{Comparison technique.}
        \label{falseNegTechnique}
    \end{subfigure}
    \hfill
    \begin{subfigure}[b]{0.74\columnwidth}
        \includegraphics[width=\textwidth]{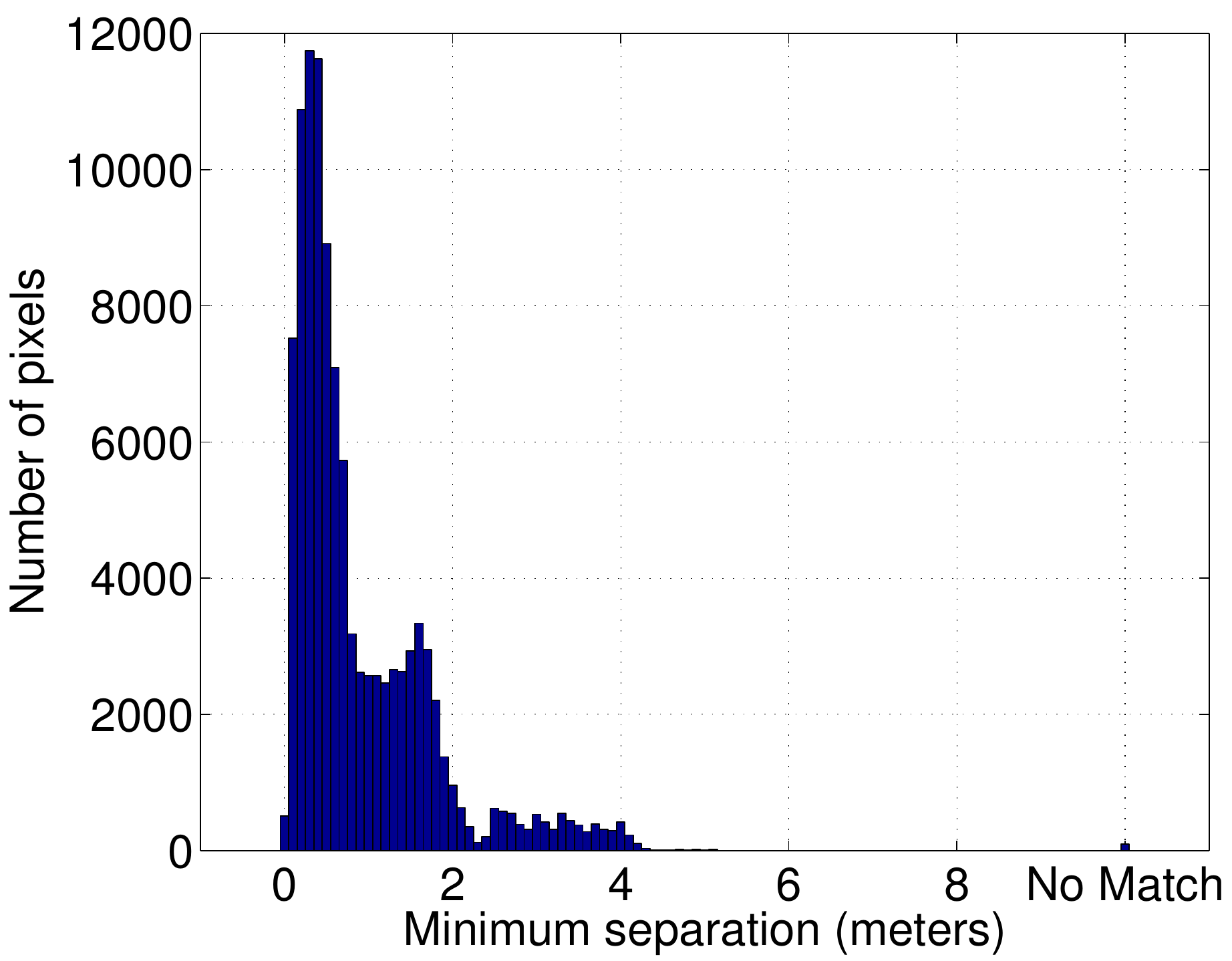}
        \caption{False-negative comparison.}
        \label{falseNegData}
    \end{subfigure}
    \caption{
        \label{bmStereoComparisonFalseNeg}
        Results of the false-negative benchmark on flight data.  In this comparison, we compute distance from
        each StereoBM point (white) to the nearest pushbroom stereo coordinate (red).  False-negatives stand
        out with large distances.  Pushbroom stereo performs well, detecting an obstacle within 2.0 meters of
        StereoBM 91.3\% of the time.
    }
\end{figure}

We compared those data to pushbroom stereo's 3D data in two ways.  First, we performed the same false-positive
detection as in Section \ref{sec:singleDisparityResults}, except we included \textit{all 3D points seen
and remembered} as we flew forward.  Second, we searched for false-negatives, or obstacles pushbroom stereo
missed, by computing the distance from each StereoBM coordinate to the nearest 3D coordinate seen and
remembered by pushbroom stereo (Figure \ref{falseNegTechnique}).

\begin{figure}
    \centering
    \includegraphics[width=.9\columnwidth]{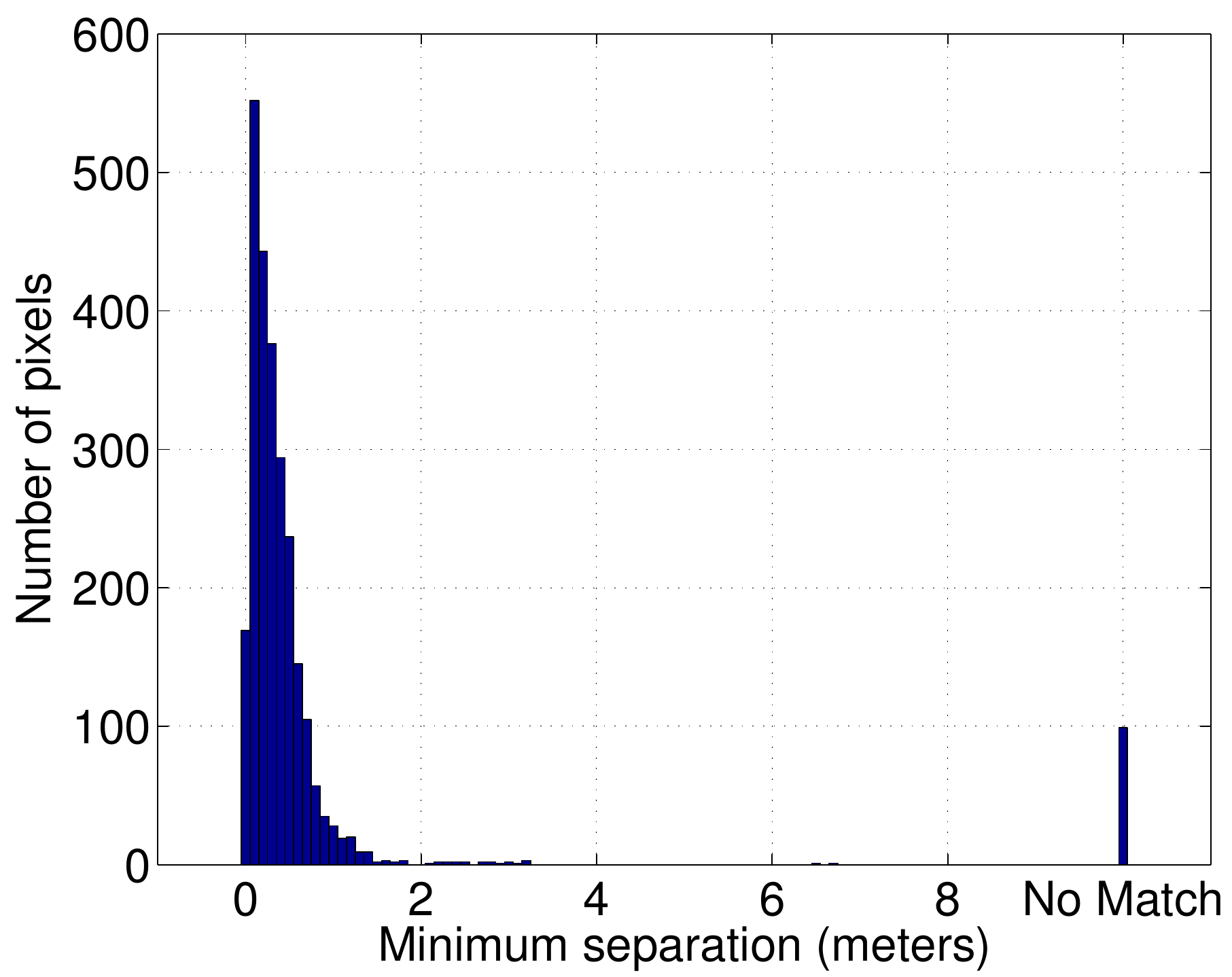}

    \caption{
        \label{bmStereoComparisonFalsePos}
        Results of the comparison as described in Figure \ref{bmStereoSketch} (a).  Our system produces few
        outliers (74.8\% and 92.6\% within 0.5 and 1.0 meters respectively), even as we integrate our state
        estimate, and the obstacle positions, forward.  \textit{No Match} indicates points that pushbroom
        stereo detected but there were no block-matching stereo detections on that frame.
    }
    \vspace{-5pt}
\end{figure}

Figures \ref{bmStereoComparisonFalsePos} and \ref{falseNegData} show the results of the
false-positive and false-negative benchmarks on all three flights respectively.  Our system does not produce
many false-positives, with 74.8\% points falling within 0.5 meters and 92.6\% falling less than one meter
from OpenCV's StereoBM implementation.  For comparison, a system producing random points at approximately the
same frequency gives approximately 1.2\% and 3.2\% for 0.5 and 1.0 meters respectively.

As Figure \ref{bmStereoComparisonFalseNeg} shows, pushbroom stereo detects most of the points on obstacles
that StereoBM sees, missing by 1.0 meters or more 32.4\% of the time.  A random system misses approximately
86\% of the time by the same metric.  For context, the closest our skilled pilot ever flew to an obstacle was
about two meters.

These metrics demonstrate that the pushbroom stereo system scarifies a limited amount of performance for a
substantial reduction in computational cost, and thus a gain in speed.  Finally, we note that all data in
this paper used identical threshold, scoring, and camera calibration parameters.

%\begin{figure}
%\begin{center}
%\includegraphics[width=\columnwidth]{figures/plots/stereo/pushbroom_to_bm_3_passes.pdf}
%\includegraphics[width=\columnwidth]{figures/plots/stereo/pushbroom_to_bm_3_passes_random.pdf}
%\caption{
    %\label{bmStereoComparison}
    %Results of the comparison as described in Figure \ref{bmStereoSketch}.  Our system produces few outliers,
    %even as we integrate our state estimate, and the obstacle positions, forward.  ``No BM Stereo" indicates
    %points that pushbroom stereo detected but there were no block-matching stereo detections on that frame.
%}
%\end{center}
%\end{figure}

%\section{Discussion}
%% why we have better data in flight
%%   more points
%%   better environment

%To successfully fly through a forest, we need to quickly detect obstacles.  We do not, however, need to see
%every point on the obstacles -- just enough to know to avoid it.  Furthermore, we make a trade-off between
%false-positives and false-negatives.  Here we have chosen to minimize false-positives at the expense of
%depth-map density.

%Based on these data, we believe that we can use this system for autonomous flight through a forest.

\section{Conclusion}

Here we describe a system that performs stereo detection with a single disparity.  A natural extension would
be to search at multiple disparities, perhaps enabling tracking of obstacles through two or more depths. As
computational power increases, we can increase the number of depths we search, continuously improving our detection.

We have demonstrated a novel stereo vision algorithm for high-framerate detections of obstacles.  Our system
is capable of quickly and accurately detecting obstacles at a single disparity and using a state-estimator to
update the position of obstacles seen in the past, building a full, local, 3D map.  It is capable of running
at 120fps on a standard mobile-CPU and is lightweight and robust enough for flight experiments on
small UAVs.  This system will allow a new class of autonomous UAVs to fly in clutter with all perception and
computation onboard.


%%%%%%%%%%%%%%%%%%%%%%%%%%%%%%%%%%%%%%%%%%%%%%%%%%%%%%%%%%%%%%%%%%%%%%%%%%%%%%%%

%%%%%%%%%%%%%%%%%%%%%%%%%%%%%%%%%%%%%%%%%%%%%%%%%%%%%%%%%%%%%%%%%%%%%%%%%%%%%%%%

%%%%%%%%%%%%%%%%%%%%%%%%%%%%%%%%%%%%%%%%%%%%%%%%%%%%%%%%%%%%%%%%%%%%%%%%%%%%%%%%
%\section*{APPENDIX}
%\section*{ACKNOWLEDGMENT}
%
%The preferred spelling of the word ÒacknowledgmentÓ in America is without an ÒeÓ after the ÒgÓ. Avoid the stilted expression, ÒOne of us (R. B. G.) thanks . . .Ó  Instead, try ÒR. B. G. thanksÓ. Put sponsor acknowledgments in the unnumbered footnote on the first page.

%%%%%%%%%%%%%%%%%%%%%%%%%%%%%%%%%%%%%%%%%%%%%%%%%%%%%%%%%%%%%%%%%%%%%%%%%%%%%%%%

%\bibliographystyle{abbrv}

%\bibliography{elib}

\end{document}